%% file: main.tex
\documentclass[conference]{IEEEtran}
\IEEEoverridecommandlockouts
\usepackage{microtype}
\usepackage{graphicx}
\usepackage{booktabs} 
\usepackage{subcaption}

\usepackage{times} 
\usepackage{balance}
\usepackage{enumitem}
\usepackage{multirow}
\usepackage{bbm}

\usepackage{tablefootnote}
\usepackage{threeparttable}
\usepackage{xspace}
\usepackage{amsmath}
\usepackage{amsfonts}
\usepackage{amssymb}
\usepackage{bm}
\usepackage{makecell}
\usepackage{setspace}
\usepackage{amsmath}
\usepackage{amssymb}
\usepackage{mathtools}
\usepackage{amsthm}
\usepackage{nomencl}
\usepackage{multirow}
\usepackage{adjustbox}
\usepackage{multirow}
\usepackage{./algorithm}
\usepackage{./algorithmic}
\def\ie{\textit{i.e.}\xspace}

\def\eg{\textit{e.g.}\xspace}

\usepackage{tikz}
\newcommand*{\circled}[1]{\lower.8ex\hbox{\tikz\draw (0pt, 0pt) circle (.47em) node {\makebox[0.4em][c]{\small #1}};}}

\begin{document}

\title{Lightweight and Post-Training Structured Pruning for On-Device Large Lanaguage Models}

\author{
\IEEEauthorblockN{
    Zihuai Xu~~Yang Xu~~Hongli Xu~~Yunming Liao~~Zhiwei Yao~~Zuan Xie
}
\IEEEauthorblockA{
    School of Computer Science and Technology, University of Science and Technology of China\\
    Suzhou Institute for Advanced Research, University of Science and Technology of China\\
    \{zihuaixu, ymliao98, zhiweiyao, xz1314\}@mail.ustc.edu.cn,~\{xuyangcs, xuhongli\}@ustc.edu.cn
}
}

\maketitle

\thispagestyle{plain} 

\begin{abstract}
\input{contents/abstract}
\end{abstract}


\section{Introduction}
\input{contents/intro}

\section{Related Work}\label{sec:relatedwork}
\input{contents/relatedwork.tex}
\section{Preliminaries and Motivation}\label{sec:prelim}
\input{contents/prelim.tex}


\section{Hybrid-granularity Model Pruning}\label{sec:algorithm}
\input{contents/overview}


\section{Experiments and Evaluation}\label{sec:evaluation}
\input{contents/simulation.tex}


\section{Conclusion}\label{sec:concluesion}
\input{contents/conclusion}

\balance
\bibliographystyle{IEEEtran}
\bibliography{contents/refs}


\end{document}

%% file: contents/abstract.tex
Considering the hardware-friendly characteristics and broad applicability, structured pruning has emerged as an efficient solution to reduce the resource demands of large language models (LLMs) on resource-constrained devices. 
Traditional structured pruning methods often need fine-tuning to recover performance loss, which incurs high memory overhead and substantial data requirements, rendering them unsuitable for on-device applications. 
Additionally, post-training structured pruning techniques typically necessitate specific activation functions or architectural modifications, thereby limiting their scope of applications. 
Herein, we introduce COMP, a lightweight post-training structured pruning method that employs a hybrid-granularity pruning strategy.  
COMP initially prunes selected model layers based on their importance at a coarse granularity, followed by fine-grained neuron pruning within the dense layers of each remaining model layer.
To more accurately evaluate neuron importance, COMP introduces a new matrix condition-based metric. 
Subsequently, COMP utilizes mask tuning to recover accuracy without the need for fine-tuning, significantly reducing memory consumption. 
Experimental results demonstrate that COMP improves performance by 6.13\% on the LLaMA-2-7B model with a 20\% pruning ratio compared to LLM-Pruner, while simultaneously reducing memory overhead by 80\%.

%% file: contents/intro.tex
Large language models (LLMs) have demonstrated remarkable success in various practical applications \cite{yi2024survey, zhuang2023toolqa, jin2024comprehensive}. 
The exceptional performance is primarily attributed to their immense number of parameters. 
However, LLMs always require substantial computing power and memory size for effective and efficient inference \cite{lin2024exploding, patterson2021carbon}, which severely impedes deployment on resource-constrained hardware, like edge modules for autonomous vehicles \cite{cui2024drive} or mobile platforms \cite{yin2024llm}.
To address these concerns, model compression has become a key solution to reduce resource footprint and/or accelerate inference speed of LLMs \cite{brown2020language, strubell2020energy}.

As one of the dominant model compression techniques, model pruning is widely used and can be categorized into unstructured pruning, structured pruning, and semi-structured pruning.  
Concretely, unstructured pruning is permitted to remove individual weights (\ie, connections) of a model without adhering to any predefined structure (\eg, neurons, attention heads) \cite{han2015learning, frankle2018lottery, frankle2019stabilizing}, and semi-structured pruning typically removes groups of weights that adhere to certain structural constraints, such as 2:4 (2 out of every 4 consecutive weights are selected and pruned) or 4:8 \cite{frantar2023sparsegpt, zhang2024plug}.
As the removed weights are set to zero, the resulting sparse weight matrices necessitate specialized hardware, \eg, NVIDIA Ampere GPUs \cite{choquette2021nvidia}, for sparse operations, potentially limiting the applicability of both unstructured and semi-structured pruning.
In contrast, structured pruning directly removes structural components such as neurons, attention heads, or entire layers \cite{kwon2022fast, zhang2024loraprune, men2024shortgpt}. 
This makes structured pruning more computationally efficient and less dependent on specialized hardware for performance gains, facilitating broader applicability across various hardware platforms.

Given its hardware-friendly nature, we opt to implement structured pruning for the deployment of LLMs on ubiquitous and resource-constrained devices.
However, existing structured pruning methods experience two potential drawbacks.
First, the gradient-based importance evaluation of structural components will result in excessive memory overhead.
For example, LLM-Pruner \cite{ma2023llmpruner} requires over 32GB memory to compute group importance when pruning a 7B model, making the process particularly resource-intensive.
Second, given the same pruning ratio, explicitly removing entire components in structured pruning generally leads to larger accuracy drops than those induced by unstructured pruning \cite{sunsimple}.
Such performance deterioration has led prior studies \cite{han2024parameter,hu2021lora,karimi2021compacter} to rely on fine-tuning to restore accuracy. 

Nevertheless, incorporating fine-tuning into model pruning also faces two new barriers in resource-constrained environments. 
On one hand, the backward pass usually doubles the memory overhead of a forward pass, while some stateful optimizers such as AdamW \cite{loshchilovdecoupled} will increase memory requirements to triple the forward pass. 
For example, LoRAPrune \cite{zhang2024loraprune} demands approximately 18GB of memory to fine-tune a 7B model, which is impractical for many devices equipped with less than 16GB of memory.
On the other hand, fine-tuning generally requires a large amount of high-quality labeled data, \eg, 50k samples in \cite{ma2023llmpruner}.
Since end devices often lack sufficient labeled data, fine-tuning must be performed on cloud servers that aggregate data from these devices.
However, the data collection process raises privacy concerns, especially when handling sensitive information like medical \cite{wiest2024privacy} or financial \cite{paul2023optimizing} data.
Consequently, there is an increasing urgency for a lightweight and on-device structured pruning technique that circumvents the need for fine-tuning.

In response, post-training structured pruning has drawn growing attention for its ability to ease both memory and data demands. 
A pioneering yet relatively straightforward method, ShortGPT \cite{men2024shortgpt}, prunes a model by outright removing highly redundant layers (identified by the strong similarities between layer outputs and inputs) without fine-tuning. 
While ShortGPT demonstrates promising results for LLaMA \cite{touvron2023llama}, its performance falls short for most other LLMs.
Meanwhile, SliceGPT \cite{ashkboos2024slicegpt} utilizes principal component analysis (PCA) to reduce matrix dimensions of models while preserving inference performance.
Besides, FPT \cite{kwon2022fast} employs mask tuning on the remaining weight matrices after model pruning by solving a least squares optimization problem, thereby requiring significantly less data compared to conventional fine-tuning.
Despite their advantages, these methods are limited in generalizability.
Specifically, SliceGPT relies on model-specific adapters, while FPT is designed for small-scale models with BERT-like architectures. 
This underscores the need for more versatile and broadly applicable post-training pruning techniques.

To this end, we propose COMP, a lightweight and widely applicable post-training structured pruning method.
COMP adopts a hybrid-granularity pruning strategy that combines layer-grained and neuron-grained pruning. 
For ease of description, we refer to model layers in an LLM as \textit{layers} and linear layers in each model layer as \textit{denses}. 
The process begins by assessing the importance of each layer, performing layer-grained pruning, and determining individual pruning ratios for the remaining layers. 
Subsequently, COMP evaluates the importance of neurons based on matrix condition and further prunes denses within each remaining layer using neuron-grained pruning. 
Inspired by FPT, COMP incorporates mask tuning to iteratively prune neurons and restore the performance of the pruned LLMs.

Our method offers two key advantages. 
First, \textit{low memory consumption}.
During layer-grained pruning, layers are dynamically loaded for importance evaluation and pruning, ensuring only a single layer is processed in memory at a time. 
Besides, neuron importance is evaluated within individual denses in neuron-grained pruning, avoiding the calculation of the entire model's gradients. 
Second, \textit{broad applicability}.
The pruning process is agnostic to model sizes and layer-internal structures, enabling broad adaptation to LLMs with diverse architectures. 
However, considering the variability in function and pruning tolerance among different denses, it is quite challenging to determine the appropriate number of neurons to prune while maintaining model performance during neuron-grained pruning.
In summary, our main contributions are as follows:
\begin{itemize}[itemsep=0pt]
\item We introduce COMP, a lightweight post-training structured pruning method.
It efficiently prunes the model with a hybrid-granularity (layer-grained and neuron-grained) pruning strategy and can handle LLMs of various structures with low memory consumption.

\item COMP enables the dynamical loading of each layer for importance evaluation and pruning, thereby saving an amount of memory overhead for model deployment.

\item We propose a key evaluation metric for determining the importance of neurons. 
Building upon the metric, we design an iterative pruning method to automatically determine the number of pruning neurons in each dense, and perform mask tuning, ensuring the dense output remains effectively unchanged after pruning.

\item 
We conduct extensive experiments on LLMs with three different structures.
Experimental results indicate that COMP can prune the LLaMA-2-7B model with only 8GB of memory, retaining approximately 91.2\% of the original performance at a 20\% pruning ratio. This represents a 6.13\% improvement compared to the state-of-the-art method LLM-Pruner.
\end{itemize}

%% file: contents/relatedwork.tex
In this study, we concentrate on structured pruning for deploying large language models (LLMs) on resource-constrained devices, leveraging its inherent hardware-friendly characteristics and broad applicability. 
Among the pioneering structured pruning techniques for LLMs, LLM-Pruner \cite{ma2023llmpruner} segments the coupled structures of an LLM into distinct groups and assesses the importance of each group through backpropagation. 
Subsequently, LLM-Pruner removes the less significant groups and fine-tunes the pruned model to restore its inference performance. 
However, the reliance on backpropagation incurs substantial memory overhead, limiting its practicality for devices with restricted memory resources. 
To address this limitation, Dejavu \cite{liu2023deja} introduces an auxiliary classifier functioning as a neighbor searcher to predict the contextual sparsity of each layer in real time. 
This classifier determines which neurons are active for a given input, thereby loading only the necessary parameters for computation. 
Additionally, LoRAPrune \cite{zhang2024loraprune} mitigates memory consumption by replacing global gradients with low-rank gradients during fine-tuning, effectively interweaving fine-tuning with pruning. 
Nonetheless, both Dejavu and LoRAPrune necessitate the training of additional neural networks and extensive fine-tuning, which require significant labeled data and further strain the memory resources of end devices.


To overcome these challenges, post-training pruning techniques have garnered increasing attention due to their reduced dependency on computational resources and labeled data. 
FPT \cite{kwon2022fast} utilizes the Fisher information matrix to explore various combinations of module masks within the model, aiming to identify the optimal submodel that minimizes the loss induced by pruning. 
It subsequently employs mask tuning to recover the performance degradation resulting from pruning. 
However, FPT's mask structure is specifically tailored for models with BERT-like architectures \cite{devlin2019bert}, thereby limiting its generalizability. 
ShortGPT \cite{men2024shortgpt} offers a more straightforward, structure-agnostic pruning approach by directly eliminating redundant model layers. 
Despite its simplicity, ShortGPT demonstrates effectiveness primarily for LLaMA \cite{touvron2023llama} and falls short when applied to other LLMs. 
In contrast, SliceGPT \cite{ashkboos2024slicegpt} exploits computational invariance within the model and employs PCA to prune dimensions in the parameter matrix with low sample variance, thereby supporting both LLaMA and OPT models. 
Nevertheless, SliceGPT's reliance on PCA means that the quantity and quality of available calibration data significantly influence its performance. 
Furthermore, SliceGPT necessitates model-specific adapters to accommodate the new inference and computational logic of the pruned model, thereby restricting its applicability across diverse models. 
In summary, existing structured pruning methods often prove unsuitable for deployment on resource-constrained local devices with limited data availability, highlighting the need for more versatile and efficient post-training pruning techniques.

%% file: contents/prelim.tex
\subsection{Preliminaries}
An LLM consists of $L$ identical layers, and each layer consists of $K$ distinct denses.
There is a weight matrix $W^{l,k} \in \mathbb{R}^{p^{l,k} \times q^{l,k}}$ and a bias vector $b^{l,k} \in \mathbb{R}^{p^{l,k}}$ at the $k$-th dense in the $l$-th layer, where $p^{l,k}$ and $q^{l,k}$ denote the output dimension and input dimension, respectively.
$X^{l,k} \in \mathbb{R}^{q^{l,k} \times T}$ denotes the inputs to the $k$-th dense in the $l$-th layer, where $T$ is the number of tokens the available calibration data.

\textbf{Structured pruning.}
Layer-grained pruning is the most simple structured pruning method.
Since Transformer-based models share the same layer structure, layer-grained pruning removes entire layers directly.
As another common pruning method, the neuron-grained pruning can be categorized into input neuron pruning and output neuron pruning, which correspond to removing a specific column or row of the parameter matrix, respectively.
Output neuron pruning reduces the output dimension of the dense layer, thereby indirectly changing the input dimension of subsequent dense layers, so we employ input neuron pruning to avoid additional impacts.
Specifically, pruning the $j$-th input neuron at the $k$-th dense in the $l$-th layer amounts to removing the $j$-th column of $W^{l,k}$ by setting it to zero, \ie, $W^{l,k}_{:,j} = \mathbf{0}$. 
This process corresponds to ignoring the $j$-th dimension of the dense's input.
Assuming $c$ neurons are pruned, the $t$-th column of the output of the $k$-th dense in the $l$-th layer can be expressed as
\begin{equation}\label{mask}
\vspace{-0.2cm}
W^{l,k}\bigl(m_c \circ X^{l,k}_t\bigr) + b^{l,k}
\end{equation}
where $\circ$ means Hadamard product, $X^{l,k}_t \in \mathbb{R}^{q^{l,k} \times 1}$ represents the $t$-th column of $X^{l,k}$, $m_c \in \{0,1\}^{q^{l,k} \times 1}$ is the input mask, and $c$ entries in $m_c$ are set to zero.

\textbf{Mask tuning.} We choose to use mask tuning to ensure the effectiveness of post-training pruning.
Mask tuning reconstructs the original dense's output using unpruned neurons. 
It reduces the discrepancy in dense outputs before and after pruning, formally written as
\begin{equation}\label{maskTuning}
\vspace{-0.2cm}
\mathop{\min}\limits_{\hat{m}_c} \mathbb{E}_{t}\Bigl[\bigl\|W^{l,k}\bigl(m_c \circ \hat{X}^{l,k}_t\bigr) \circ \hat{m}_c - W^{l,k}X^{l,k}_t\bigr\|_2^2\Bigr]
\end{equation}
where $\hat{m}_c$ is the tuned mask.
The input to the dense layer is different between the pruned model and unpruned model due to the influence of preceding layers, so we use $\hat{X}^{l,k}$ and $X^{l,k}$ as the inputs individually.

\begin{figure}[t]
\centering
\includegraphics[width=.85\linewidth]{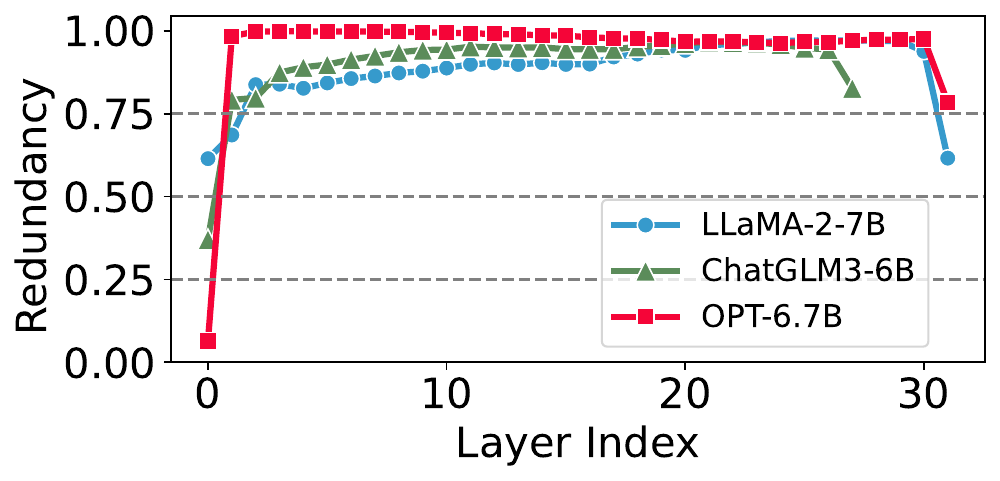}
\vspace{-0.4cm}
\caption{Layer redundancy of different models.}
\label{fig:LayerRedundancy}
\vspace{-0.5cm}
\end{figure}


\begin{figure*}[t]
\centering
\begin{subfigure}{.32\textwidth}
\centering
\includegraphics[height=3.56cm]{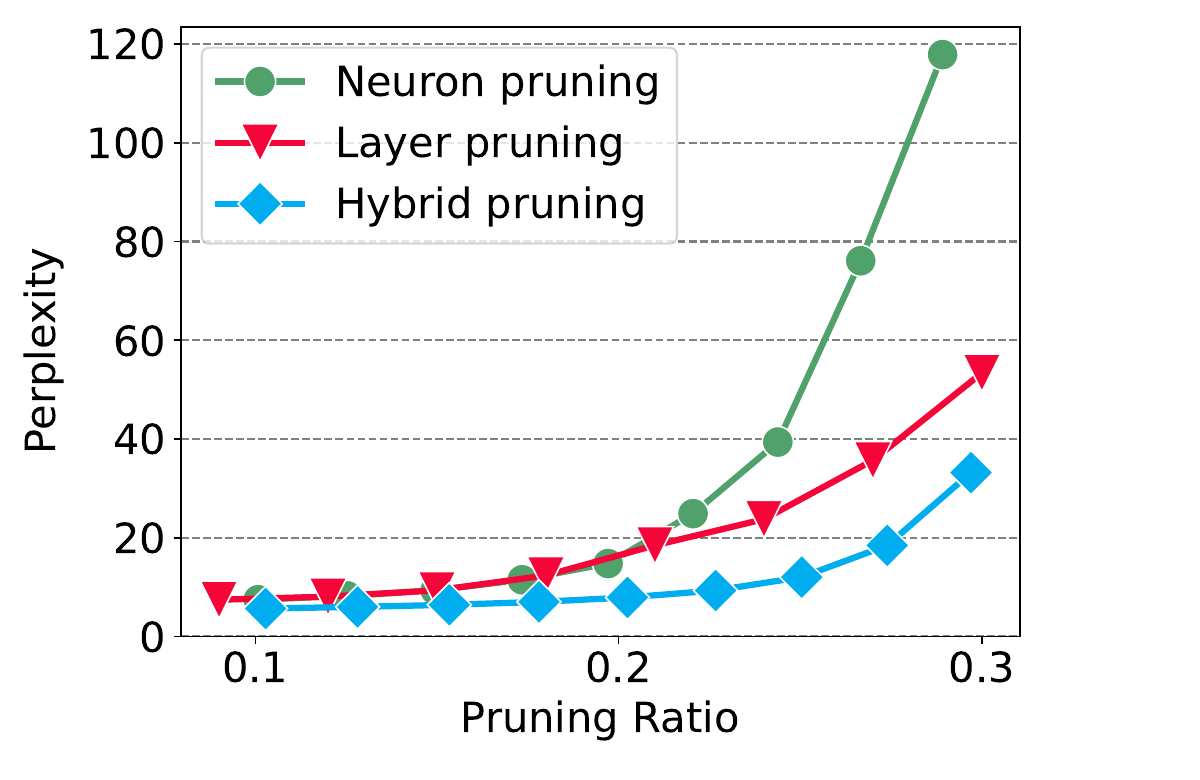}
\caption{LLaMA-2-7B}
\label{fig:PPL_Pruneratio1}
\end{subfigure}
\begin{subfigure}{.32\textwidth}
\centering
\includegraphics[height=3.5cm]{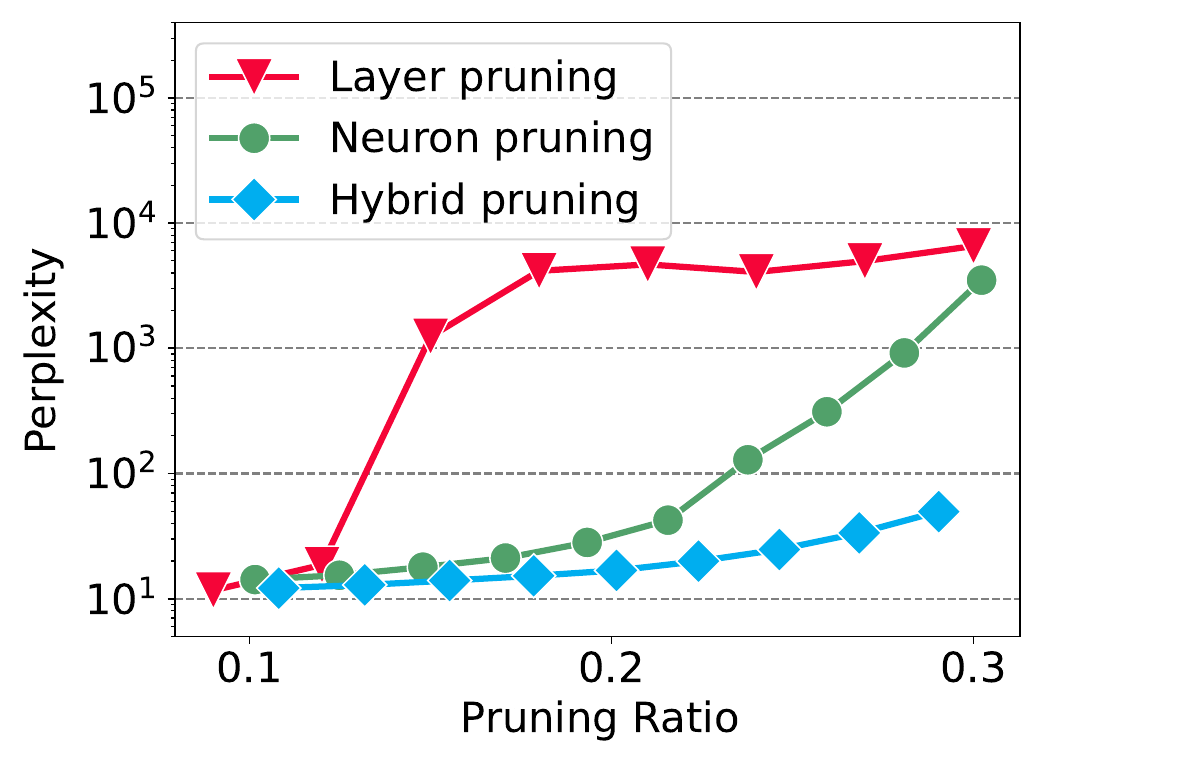}
\caption{OPT-6.7B}
\label{fig:PPL_Pruneratio3}
\end{subfigure}
\begin{subfigure}{.32\textwidth}
\centering
\includegraphics[height=3.5cm]{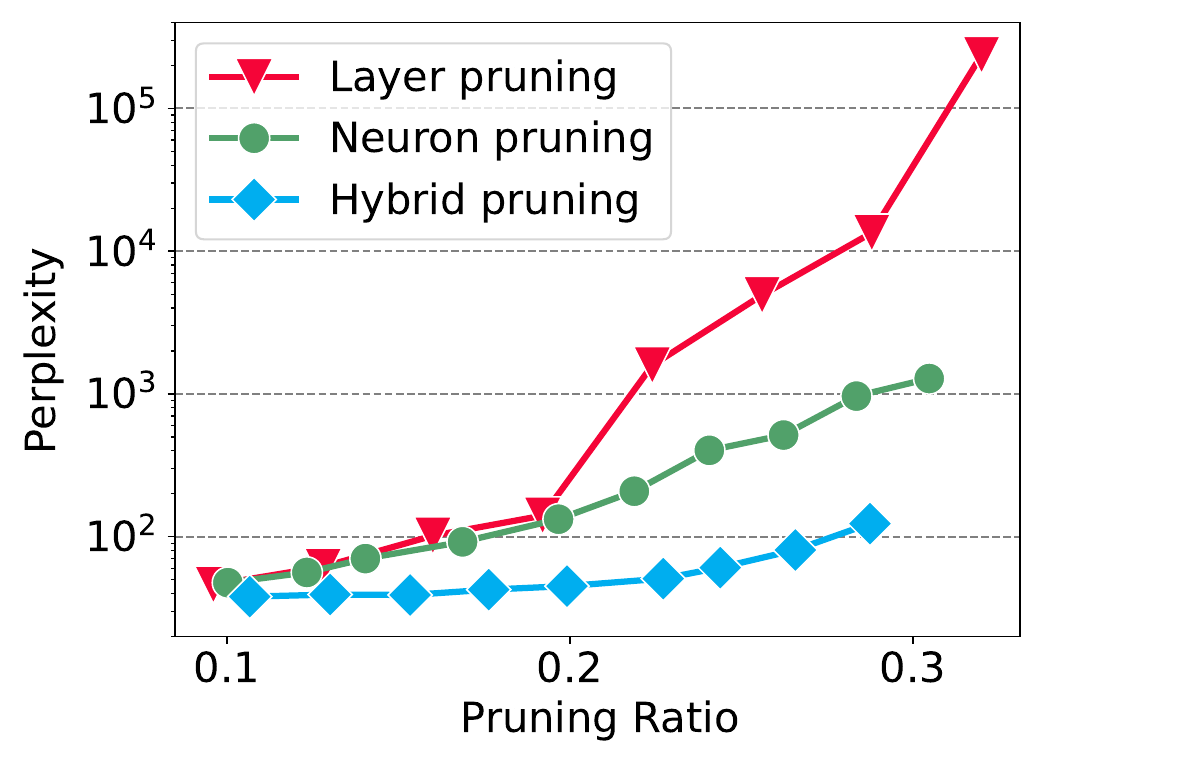}
\caption{ChatGLM3-6B}
\label{fig:PPL_Pruneratio2}
\end{subfigure}
\caption{Wikitext2 perplexity of different models with three strategies under certain pruning ratios.}
\vspace{-0.4cm}
\label{fig:PPL_Pruneratio}
\end{figure*}

\subsection{Intuition of Metric Design}
\textbf{Layer Importance.}
In the OPT model, Dejavu \cite{liu2023deja} observes layer redundancy, characterized by the high similarity between each layer's output and input.
Intuitively, if the redundancy of a layer is high, this layer will not significantly impact the final result and can be directly skipped during inference. 
Inspired by this, layer redundancy is defined as the cosine similarity between the layer's output and input.
We test the redundancy of each layer in different models, and the results are shown in Figure \ref{fig:LayerRedundancy}.
Obviously, most layers in LLMs exhibit high redundancy, with the redundancy being low in the first and last layers and high in the middle layers.
In other words, layers with higher redundancy should have lower layer importance. 
Therefore, the importance of the $l$-th layer is defined as follows:
\begin{equation}\label{layerimp}
I_l = 1 - \mathbb{E}_{t}\Bigl[\frac{(X^l_{t})^TX^{l+1}_t}{||X^{l}_t||_2||X^{l+1}_t||_2}\Bigr]
\end{equation}
where the similarity term represents the layer redundancy of the $l$-th layer, $X^l$ and $X^{l+1}$ represent the input and output of the $l$-th layer respectively. 
Note that Eq. \eqref{layerimp} only involves the parameters of the $l$-th layer, so the layers of the model can be dynamically loaded to reduce overhead.

\textbf{Neuron Importance.}
When pruning neurons, we expect to prune some neurons that have a slight impact on the final result. 
As shown in Eq. \eqref{maskTuning}, the $k$-th dense unit in the $l$-th layer is expected to reconstruct the original output more accurately after mask tuning.
In other words, neuron pruning should contribute to mask tuning; therefore, neuron importance is designed based on the optimization goal.
Eq. \eqref{maskTuning} can be further expressed as:
\begin{equation}\label{lsq}
\vspace{-0.2cm}
\mathop{\min}\limits_{\hat{m}_c} \mathbb{E}_{t}\Bigl[\bigl\| \bigl(W^{l,k} \mathrm{Diag}(m_c \circ \hat{X}^{l,k}_t)\bigr)\hat{m}_c - W^{l,k}X^{l,k}_t\bigr\|_2^2\Bigr]
\end{equation}
where $\mathrm{diag}(x)$ means the diagonal matrix of vector $x$.
Eq. \eqref{lsq} is actually a least squares problem that can indeed be rewritten as a standard linear equation:
\begin{equation}
\label{lsq1}
\vspace{-0.2cm}
\hat{A}^T\hat{A}\hat{m}_c = \hat{A}^Ty
\end{equation}

where $\hat{A} = \mathbb{E}_{t}[W^{l,k} \mathrm{Diag}(m_c \circ \hat{X}^{l,k}_t)]$, $y=\mathbb{E}_{t}[W^{l,k}X^{l,k}_t]$.
It is natural to focus on the properties of the coefficient matrix $\hat{A}^T\hat{A}$. 
As stated in \cite{cline1979estimate}, the matrix condition number is a commonly used metric to measure the properties of the coefficient matrix. 
For a linear equation system $\hat{A}^T\hat{A}m_c = \hat{A}^Ty$, the condition number of the coefficient matrix $\kappa(\hat{A}^T\hat{A})$ is defined as:
\begin{equation}\label{condA}
\kappa(\hat{A}^T\hat{A}) = ||\hat{A}^T\hat{A}||_2 \cdot ||(\hat{A}^T\hat{A})^{-1}||_2
\end{equation}
where $||\hat{A}^T\hat{A}||_2$, known as the spectral norm, denotes the maximum singular value of $\hat{A}^T\hat{A}$.
The condition number indicates the sensitivity of the computation to errors.
Concretely, a large condition number of the coefficient matrix $\kappa(\hat{A}^T\hat{A})$ suggests that this matrix is poorly conditioned, and solving for mask $\hat{m}_c$ might lead to significant numerical errors.
In mask tuning, the pruned model is expected to handle most tasks effectively, meaning that the solution $\hat{m}_c$ can minimize the optimization objective in most cases.
In other words, $\kappa(\hat{A}^T\hat{A})$ should be reduced to minimize the errors of $\hat{m}_c$.

Note that $\kappa(\hat{A}^T\hat{A})$ can be approximated by the second-order of Taylor expansion around initial mask $\mathbf{1}$:
\begin{align}
\kappa(\hat{A}^T\hat{A}) &\approx \kappa(A^TA) - \mathbf{g}^{T}(\mathbf{1} - m_c)\nonumber\\
&+ \frac{1}{2} (\mathbf{1} - m_c)^{T}\mathbf{H}(\mathbf{1} - m_c) 
\end{align}
where $\mathbf{g} = \frac{\partial}{\partial m_c} \kappa(A^TA)$, $ \mathbf{H} = \frac{\partial^2}{\partial^2 m_c} \kappa(A^TA)$, and $A=\mathbb{E}_{t}\Bigl[W^{l,k} \mathrm{Diag}(\hat{X}^{l,k}_t)\Bigr]$ is the initial coefficient matrix before pruning.
Since forming the exact Hessian matrix explicitly is infeasible, $\mathbf{H}$ can be approximated with the (empirical) Fisher information matrix $\mathbf{F}$ of the mask variables as:
\begin{align}
\mathbf{F} = (\frac{\partial}{\partial m_c} \kappa(A^TA))(\frac{\partial}{\partial m_c} \kappa(A^TA))^{T}
\end{align}
It is also hard to use the full Fisher information matrix $\mathbf{I}$.
Therefore, the input neurons are assumed to be independent, meaning that $\mathbf{F}$ is diagonal.
So the importance of the $f$-th neuron in a certain dense is defined as follows:
\begin{align}\label{neuronimp}
    \widetilde{I}_f &= -\mathbf{g}_f + \frac{1}{2} \mathbf{F}_{ff} \nonumber\\
    &= -[\frac{\partial}{\partial m_c} \kappa(A^TA)_a]_{f} + \frac{1}{2} [\frac{\partial}{\partial m_c} \kappa(A^TA)]^2_{f}
\end{align}
Pruning neurons with higher importance clearly results in a more significant increase in the condition number of the coefficient matrix, which in turn makes the solution more sensitive to different inputs of the dense, potentially leading to instability.

\begin{figure*}[t]
\centering
\includegraphics[width=0.95\textwidth]{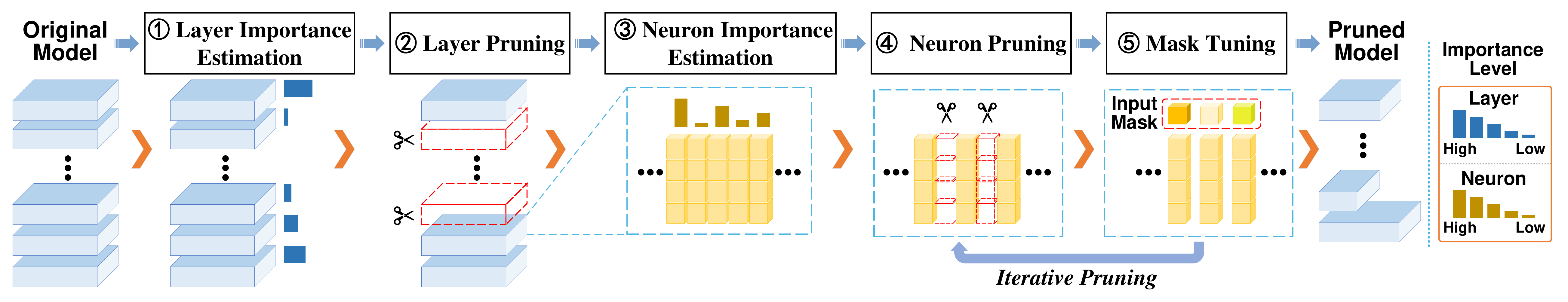}
\caption{Illustration of the pruning process of COMP.}
\label{fig:overview}
\vspace{-0.3cm}
\end{figure*}

Solving $\kappa(A^TA)$ in Eq. \eqref{neuronimp} involves inverting $A^TA$, which is not necessarily invertible.
Noticing that $A^TA$ is a positive semidefinite symmetric matrix, we add $\epsilon E$ to $A^TA$, where $E$ is the identity matrix and $\epsilon$ is a very small positive value.
Since the modified matrix is positive definite, the process of calculating its inverse can be accelerated by the Cholesky decomposition \cite{krishnamoorthy2013matrix}.


\subsection{Motivation}
We conduct tests on three different pruning strategies over different models: (1) \textit{Layer pruning} only removes layers.
(2) \textit{Neuron pruning} only removes the same number of neurons from each dense.
(3) \textit{Hybrid pruning} removes 2 layers and then removes the same number of neurons from each dense.
We use Wikitext2 perplexity to represent model performance, and the results are shown in Figure \ref{fig:PPL_Pruneratio}.

We give two conclusions from these results. First, pruning a small number of layers will not greatly impact the model performance. 
For example, the perplexity only increases slightly from 5.42 to 7.43 when removing 3 layers in LLaMA-2-7B by Figure \ref{fig:PPL_Pruneratio}(a).
However, merely applying layer pruning leads to a substantial decrease in performance once a certain pruning ratio is reached. 
For example, the perplexity of OPT-6.7B rises from 18.64 to over 1200 when the number of removed layers increases from 4 to 5 in Figure \ref{fig:PPL_Pruneratio}(b). 
Second, the hybrid pruning method delivers the lowest perplexity among three pruning methods, and this advantage is obvious when the pruning ratio is high.
Specifically, in Figure \ref{fig:PPL_Pruneratio}(c), at approximately a 30\% pruning ratio for ChatGLM3-6B, the hybrid pruning method decreases the perplexity by about 90\% and 99\% compared with neuron pruning and layer pruning, respectively.

Despite achieving the highest performance, hybrid pruning's approach lacks neuron selectivity.
It is quite challenging to determine the appropriate number of neurons to prune while maintaining model performance during neuron-grained pruning.
We will design an iterative pruning process to address the challenge in the subsequent section.

%% file: contents/overview.tex
\begin{algorithm}[h]
   \caption{Pruning process of COMP.}
   \label{alg1}
\begin{algorithmic}
    \STATE {\bfseries Input:} Model $M$, Calibration data batch $X$, Pruning ratio $r$, Number of removed layers $n$.
    \STATE {\bfseries Output:} Pruned model $M'$.
    \STATE // \textit{Layer-grained Pruning}
    \REPEAT
    \STATE Calculate layer importance by Eq. \eqref{layerimp}.
    \STATE Remove the least important layer.
    \UNTIL {$n$ layers are removed.}
    \STATE // \textit{Neuron-grained Pruning}
    \FOR{Each remaining layer $l$}
    \STATE Calculate pruning ratio $r_l$ by Eq. \eqref{weight}.
    \STATE Set the variance threshold $v_T$ to 0.
    \REPEAT
    \FOR{Each dense $k$ in the $l$-th layer}
    \STATE Calculate neuron importance by Eq. \eqref{neuronimp}.
    \STATE Set the number of pruning neurons $c$ to 0.
    \REPEAT 
    \STATE Prune $c$ least important neurons.
    \STATE Tune the mask $\hat{m}_c$ by Eq. \eqref{lsq1}.
    \STATE Increase $c$.
    \UNTIL {$\text{Var}(\hat{m}_c) \geq v_T$}
    \STATE Update the number of pruned parameters. 
    \ENDFOR
    \STATE Increase $v_T$.
    \UNTIL {The pruning ratio of the $l$-th layer reaches $r_l$.}
    \ENDFOR
\end{algorithmic}
\end{algorithm}

The pruning process of COMP is illustrated in Figure \ref{fig:overview}. 
To prune an LLM, we begin by evaluating the importance of each layer (\circled{1}). 
Following this, layer-grained pruning (\circled{2}) is performed to gradually remove some less important layers. 
For the remaining layers, specific pruning ratios are determined based on their evaluated importance. 
Next, we assess the importance of neurons within the denses of each remaining layer (\circled{3}) and execute neuron-grained pruning accordingly (\circled{4}). 
Finally, mask tuning (\circled{5}) is conducted to restore model performance. 
To achieve a high pruning ratio while ensuring satisfactory model performance, neuron-grained pruning and mask tuning are performed iteratively. 
The detailed process is outlined in Algorithm \ref{alg1}.

\subsection{Layer-grained Pruning}
\label{Layer-grained Pruning}
Since the layer importance can be obtained through Eq. \eqref{layerimp}, the least important layers can be directly removed for layer-grained pruning.
However, calculating a layer's importance depends on the output of its previous layer.
If a layer is removed, the input of subsequent layers changes, thus the layer importance order derived from a single calculation may be inaccurate. 
Based on this, the layer importance order is established by iteratively calculating layer importance.
Specifically, the layer with the smallest importance is removed in each iteration, and the layer's subsequent layers are reordered based on their new importance.

\subsection{Neuron-grained Pruning}
\label{Neuron-grained Pruning}
The first step in neuron-grained pruning is to specify the pruning ratio for the remainig layers. 
Intuitively, layers with lower importance will have higher pruning ratios.
With a target pruning ratio of $r$, the pruning ratio for the $l$-th layer is defined as
\begin{align}\label{weight}
    r_l= \frac{w_l \cdot (rN - n\hat{N})}{\hat{N}}
\end{align}
where $n$ is the number of removed layers, $w_l$ represents the harmonic mean of the rest layer's importance, and $N$, $\hat{N}$ are the parameters' numbers of the model and the $l$-th layer individually.
After $r_l$ is determined, we need to determine the number of pruning neurons $c^{l,k}$ in the $k$-th dense of the $l$-th layer.
In order to find the appropriate number of pruning neurons in each dense, we adapt iterative pruning, gradually increasing $c^{l,k}$ to ensure that the output of the model changes within a controllable range.

\textbf{Iterative Pruning.}
Assume that $c$ neurons are pruned in a dense, and the corresponding pruning mask is $m_c$. 
After pruning, the tuned mask $\hat{m}_c$ is obtained through Eq. \eqref{lsq1} to reconstruct the output of the dense. 
When $c$ increases, the freedom degrees of $\hat{m}_c$ are reduced, which means that the non-zero elements in $\hat{m}_c$ have to bear more burden. 
Therefore, the change of non-zero elements will tend to be uncertain as $c$ increases, and the variance of non-zero elements in the tuned mask becomes larger. 
The large variance of mask is not conducive to the generalization of the model \cite{lever2016points}, therefore it is crucial to control the variance of non-zero elements $\text{Var}(\hat{m}_c)$.
However, the exact variance threshold cannot be calculated directly. 
To obtain a suitable threshold $v_T$, we set its initial value to zero and iteratively increase it while pruning. 
When $c$ gradually increases until $\text{Var}(\hat{m}_c)$ reaches the current threshold $v_T$ in a dense, its next dense starts to be pruned. 
If a layer has reached the target pruning ratio after all its denses are pruned, its next layer starts pruning. 
If not, increase the variance threshold $v_T$ and start pruning the layer again.

\textbf{Cumulative Deviation.}
The input of subsequent layers will change after previous layers are pruned.
During the pruning process, the errors in each layer may accumulate.
Because mask tuning aims to keep the dense outputs as close as possible to those of the original model, taking into account the impact of already-pruned layers can lead to cumulative errors, potentially causing the pruned model to overfit the calibration data.
To alleviate the cumulative deviation from the original model, we let each pruned layer fully ``absorb'' the impact of pruning within itself.
Concretely, we perform pruning a layer using the input from the original model.

\begin{table*}[t]
\centering
\caption{Zero-shot Performance of the compressed LLaMA2 models.}
\label{tab:LLaMA-2 performance}
\adjustbox{max width=\textwidth}{
\begin{tabular}{cclcccccccccc}
    \toprule
    \multirow{2}{*}{Model} & \multirow{2}{*}{Pruning Ratio} & \multirow{2}{*}{Method} & \multirow{2}{*}{WikiText2 $\downarrow$} & \multirow{2}{*}{PTB $\downarrow$} & \multirow{2}{*}{Alpaca $\downarrow$} & \multicolumn{6}{c}{Tasks} & \multirow{2}{*}{Average $\uparrow$} \\
    \cmidrule{7-12}
    & & & & & & WinoGrande & BoolQ & LogiQA & MMLU & PIQA & SCIQ & \\
    \midrule
    \multirow{13}{*}{LLaMA-2-7B} & \multirow{1}{*}{0\%} & Origin & 5.47 & 24.09 & 2.70 & 69.22 & 77.71 & 25.65 & 41.88 & 78.07 & 93.90 & 64.41 \\
    \cmidrule(lr){2-13}
    & \multirow{4}{*}{\shortstack[l]{20\%}} & 
    SLICEGPT & 777.57 & 1774.94 & 338.78 & 49.49 & 37.83 & 19.66 & 22.95 & 53.10 & 21.10 & 34.02 \\
    & & LLM-Pruner & 21.02 & 119.79 & 8.12 & 62.43 & 67.37 & 20.28 & 25.26 & \textbf{73.88} & 84.40 & 55.60 \\
    & & ShortGPT & 18.45 & 64.67 & 5.13 & 66.14 & 62.17 & 22.89 & 24.38 & 69.26 & 82.70 & 54.59 \\
    & & COMP & \textbf{10.39} & \textbf{42.18} & \textbf{3.85} & \textbf{67.09} & \textbf{71.13} & \textbf{25.81} & \textbf{28.47} & 70.84 & \textbf{89.10} & \textbf{58.74} \\
    \cmidrule(lr){2-13}
    & \multirow{4}{*}{\shortstack[l]{25\%}} &  	
    SLICEGPT & 1138.59 & 2804.58 & 671.28 & 50.36 & 37.83 & 19.97 & 23.28 & 54.24 & 22.20 & 34.65 \\
    & & LLM-Pruner & 35.97 & 182.14 & 11.58 & 63.22 & 64.62 & 21.20 & 24.78 & \textbf{71.82} & 79.60 & 54.21 \\
    & & ShortGPT & 35.68 & 98.26 & 7.28 & \textbf{64.09} & 62.17 & \textbf{25.04} & \textbf{38.37} & 65.02 & 70.10 & 54.13 \\
    & & COMP & \textbf{13.89} & \textbf{58.73} & \textbf{4.87} & 63.69 & \textbf{66.42} & 24.12 & 23.02 & 67.90 & \textbf{86.40} & \textbf{55.26} \\
    \cmidrule(lr){2-13}
    & \multirow{4}{*}{\shortstack[l]{30\%}} & 
    SLICEGPT & 1206.38 & 2562.20 & 646.90 & 50.75 & 37.83 & 19.66 & 22.95 & 52.72 & 21.40 & 34.22 \\
    & & LLM-Pruner & 47.99 & 250.99 & 18.50 & 62.12 & 61.62 & 21.35 & 23.31 & \textbf{68.88} & 72.10 & 51.56\\
    & & ShortGPT & 65.89 & 143.52 & 10.79 & \textbf{63.06} & 58.30 & \textbf{22.43} & \textbf{34.48} & 61.92 & 58.30 & 50.40 \\
    & & COMP & \textbf{19.61} & \textbf{80.83} & \textbf{6.26} & 62.43 & \textbf{63.18} & 21.66 & 24.63 & 65.56 & \textbf{81.30} & \textbf{53.13} \\
    \midrule
    \midrule
    \multirow{13}{*}{LLaMA-2-13B} & \multirow{1}{*}{0\%} & Origin & 4.88 & 34.42 & 2.62 & 72.22 & 80.61 & 26.27 & 52.07 & 79.11 & 94.60 & 67.48 \\
    \cmidrule(lr){2-13}
    & \multirow{4}{*}{\shortstack[l]{20\%}} &  	
    SLICEGPT & 929.68 & 929.68 & 433.51 & 50.12 & 37.83 & 18.89 & 22.95 & 52.29 & 20.60 & 33.78 \\
    & & LLM-Pruner & 12.39 & 86.87 & 5.72 & 68.51 & 64.25 & 22.58 & 36.01 & \textbf{76.39} & 91.40 & 59.86 \\
    & & ShortGPT & 15.50 & \textbf{50.76} & 5.44 & \textbf{71.43} & \textbf{63.21} & 24.58 & \textbf{49.40} & 73.01 & 87.30 & 61.49 \\
    & & COMP & \textbf{7.54} & 54.22 & \textbf{3.22} & 70.01 & 62.39 & \textbf{27.34} & 46.03 & 74.21 & \textbf{91.30} & \textbf{61.88} \\
    \cmidrule(lr){2-13}
    & \multirow{4}{*}{\shortstack[l]{25\%}} &  	
    SLICEGPT & 1072.27 & 1756.10 & 717.66 & 50.91 & 37.83 & 21.51 & 22.95 & 52.12 & 20.60 & 34.32 \\
    & & LLM-Pruner & 15.20 & 109.28 & 7.14 & 66.69 & 58.96 & 22.27 & 26.32 & \textbf{74.54} & 87.50 & 56.05 \\
    & & ShortGPT & 35.42 & 74.79 & 5.52 & 67.40 & 38.35 & 23.20 & \textbf{47.69} & 69.31 & 77.40 & 53.89 \\
    & & COMP & \textbf{8.94} & \textbf{69.70} & \textbf{3.57} & \textbf{69.38} & \textbf{62.32} & \textbf{25.96} & 43.23 & 73.07 & \textbf{90.00} & \textbf{60.66} \\
    \cmidrule(lr){2-13}
    & \multirow{4}{*}{\shortstack[l]{30\%}} &  	
    SLICEGPT & 1216.57 & 2636.22 & 671.22 & 51.54 & 37.83 & 20.28 & 22.95 & 51.90 & 18.20 & 33.78 \\
    & & LLM-Pruner & 22.68 & 139.52 & 9.27 & 67.40 & \textbf{64.19} & 19.82 & 25.10 & \textbf{72.74} & 85.80 & 55.84 \\
    & & ShortGPT & 50.05 & \textbf{92.10} & 6.34 & \textbf{68.43} & 37.68 & 21.97 & \textbf{47.69} & 67.46 & 74.60 & 52.97 \\
    & & COMP & \textbf{11.01} & 105.40 & \textbf{3.94} & 68.27 & 62.45 & \textbf{26.42} & 34.18 & 68.88 & \textbf{89.20} & \textbf{58.23} \\
    \bottomrule
\end{tabular}}
\vspace{-0.3cm}
\end{table*}%

%% file: contents/simulation.tex
\subsection{Experimental Settings}
\textbf{Large Language Models.}
To showcase the effectiveness and versatility of COMP, we perform tests over five open-source large language models with three kinds of structures: LLaMA-2-7B, LLaMA-2-13B \cite{touvron2023llama}, OPT-6.7B,  OPT-13B \cite{zhang2022opt}, and ChatGLM3-6B \cite{zeng2022glm}.
We compare COMP with 3 state-of-the-art algorithms for structured pruning: LLM-Pruner \cite{ma2023llmpruner}, SliceGPT \cite{ashkboos2024slicegpt} and ShortGPT \cite{ashkboos2024slicegpt}, respectively.

\textbf{Metrics and Datasets.}
Following SliceGPT \cite{ashkboos2024slicegpt}, we assess the perplexity of all models on 3 typical datasets: WikiText2 \cite{merity2016pointer}, PTB \cite{marcus1993building}, and Alpaca \cite{noach2020compressing}. 
To test the model's task-agnostic capability, we complement the evaluation of the accuracy metric with LLaMA-2-7B and LLaMA-2-13B on 6 common reasoning datasets: BoolQ \cite{clark-etal-2019-boolq}, WinoGrande \cite{ai2:winogrande}, LogiQA \cite{liu2020logiqa}, MMLU \cite{hendrycks2020measuring}, PIQA \cite{bisk2020piqa} and SCIQ \cite{SciQ}.
We select pruning rates of 20\%, 25\% and 30\%, \ie, removing 20\% to 30\% model parameters.

\textbf{Implementation Details.}
We adopt the same settings as those in LLM-Pruner.
First, we use only 10 randomly selected samples from C4 \cite{raffel2020exploring} as the calibration samples, each truncated to a sequence length of 128.
Second, we keep the first two layers and the last layer of the model free from neuron pruning.
Besides, we solve the linear equation by the LSMR solver in CuPy \cite{nishino2017cupy} to address the problem of the large size of the coefficient matrix. 
All experiments are conducted on NVIDIA A6000, with a GPU memory size of 48GB.

\begin{table*}[t]
\centering
\vspace{-0.2cm}
\caption{Perplexity results of compressed OPT-6.7B, OPT-13B and ChatGLM3-6B.}
\label{table:ppl_results}
\adjustbox{max width=.912\textwidth}{
\begin{tabular}{@{}clccccccccc@{}}
\toprule
\multirow{2}{*}{Pruning Ratio} & \multirow{2}{*}{Method} & \multicolumn{3}{c}{OPT-6.7B} & \multicolumn{3}{c}{OPT-13B} & \multicolumn{3}{c}{ChatGLM3-6B}\\
\cmidrule(lr){3-5} \cmidrule(lr){6-8} \cmidrule(lr){9-11}
& & Wikitext2 & PTB & Alpaca & Wikitext2 & PTB & Alpaca & Wikitext2 & PTB & Alpaca \\
\midrule
0\% & Origin & 10.86 & 12.17 & 3.78 & 10.13 & 11.47 & 3.68 & 29.96 & 50.31 & 6.06 \\
\midrule
\multirow{4}{*}{20\%} & SliceGPT & 2059.10 & 1879.92 & 372.91 & 10869.87 & 4887.74 & 2370.36 & $-$ & $-$ & $-$ \\
& LLM-Pruner & 31.08 & 37.08 & 8.39 & 24.71 & 27.65 & 7.11 & $-$ & $-$ & $-$ \\
& ShortGPT & 4680.30 & 3307.72 & 1583.22 & 11573.50 & 7528.65 & 4717.45 & 760.58 & 617.75 & 69.07 \\
& COMP & \textbf{19.86} & \textbf{20.45} & \textbf{5.17} & \textbf{20.10} & \textbf{23.64} & \textbf{5.30} & \textbf{60.44} & \textbf{77.86} & \textbf{8.07} \\
\midrule
\multirow{4}{*}{25\%} & SliceGPT & 4066.31 & 3198.35 & 590.36 & 7815.31 & 5871.41 & 1434.56 & $-$ & $-$ & $-$ \\
& LLM-Pruner & 53.87 & 75.87 & 14.11 & 37.53 & \textbf{38.06} & 10.25 & $-$ & $-$ & $-$ \\
& ShortGPT & 4974.92 & 3845.85 & 2183.15 & 13085.26 & 6666.44 & 3726.09 & 1417.31 & 1120.79 & 125.42 \\ 
& COMP & \textbf{31.48} & \textbf{29.56} & \textbf{6.54} & \textbf{30.10} & 38.81 & \textbf{6.87} & \textbf{79.59} & \textbf{96.92} & \textbf{10.17} \\ 
\midrule
\multirow{4}{*}{30\%} & SliceGPT & 4224.52 & 3567.74 & 982.67 & 10427.82 & 7748.62 & 2685.97 & $-$ & $-$ & $-$ \\
& LLM-Pruner & 100.06 & 293.33 & 26.41 & 74.87 & \textbf{61.90} & 15.37 & $-$ & $-$ & $-$ \\
& ShortGPT & 6558.42 & 4843.02 & 2556.63 & 11807.03 & 5925.62 & 4297.03 & 4404.26 & 3528.29 & 419.91 \\
& COMP & \textbf{82.44} & \textbf{64.97} & \textbf{11.29} & \textbf{54.35} & 67.23 & \textbf{9.45} & \textbf{140.45} & \textbf{165.37} & \textbf{24.69} \\
\bottomrule 
\end{tabular}}
\vspace{-0.3cm}
\end{table*}

\subsection{Main Results}
\textbf{Zero-shot performance.} 
Firstly, we conduct a comprehensive test on LLaMA-2-7B and LLaMA-2-13B models to verify the superiority of COMP, and the zero-shot results are shown in Table \ref{tab:LLaMA-2 performance}.
The COMP method outperforms the baseline approaches on the LLaMA2 models, particularly regarding perplexity. 
For instance, when the pruning ratio of the LLaMA-2-7B model is 30\%, COMP achieves a perplexity of 19.61 on the WikiText2 dataset, significantly better than the perplexity of 47.99 achieved by LLM-Pruner. 
We observe that SliceGPT performs poorly, and the reason is limited calibration data and the large embedding dimension of the model.
The PCA process in SliceGPT relies on the covariance matrix to derive the principal components.
When the number of data is small and the dimension is large, the estimation of the sample covariance matrix can be highly inaccurate. 
In contrast, ShortGPT exhibits commendable performance on the LLaMA2 models. 
Specifically, when pruning the 13B model, its average accuracy score (61.49) is nearly comparable to that of COMP (61.88). 
Nevertheless, as the pruning ratio escalates, ShortGPT's performance progressively declines.
On the LLaMA-2-7B model, COMP retains 91.2\%, 85.8\%, and 82.5\% of the unpruned model's performance under three pruning ratios. 
For the LLaMA-2-13B model, these retention rates rise to 91.7\%, 89.9\%, and 86.3\%.
The demonstration proves the feasibility of COMP to effectively compress the large language model, even without relying on training data.

Secondly, we test the perplexity performance of OPT-6.7B, OPT-13B, and ChatGLM3-6B to verify the versatility of COMP. 
COMP consistently outperforms other methods on the OPT models, especially on the Alpaca dataset.
In Table \ref{table:ppl_results}, given the pruning ratio of 30\%, the perplexity of COMP on the 6.7B model and 13B model is 11.29 and 9.45 respectively, which is 42.75\% and 38.52\% lower than that of LLM-Pruner.
Similar to SliceGPT, ShortGPT performs poorly on the three models, suggesting that blindly removing layers for model compression does not always work. 
Besides, both SliceGPT and LLM-pruner are unable to efficiently prune the ChatGLM3-6B model due to the inherent limitations of their methods. 
In contrast, COMP is still applicable. 
At a 20\% pruning ratio, the perplexity of COMP on ChatGLM3-6B is only about doubled compared to the original model, which is within a reasonable acceptable range.
COMP remains applicable across different models, demonstrating its versatility and adaptability.

\begin{figure}[t]
\hspace{0.02\linewidth}
\begin{subfigure}{0.43\linewidth} 
\centering
\includegraphics[height=3.5cm]{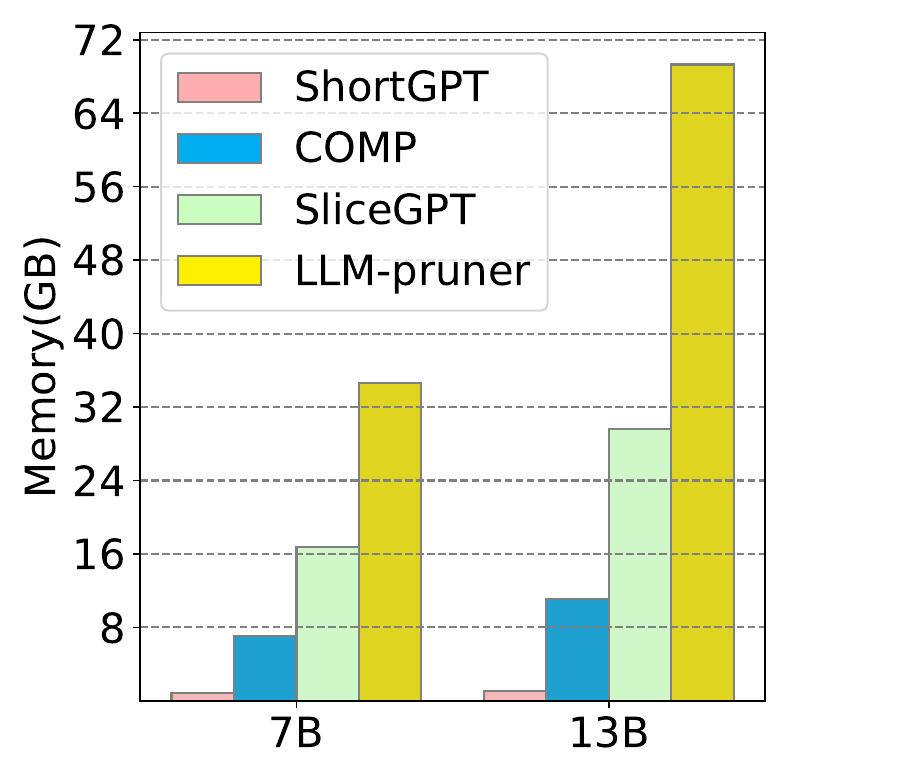}
\caption{LLaMA2}
\label{fig:bar_LLaMA}
\end{subfigure}
\hspace{0.03\linewidth}
\begin{subfigure}{0.43\linewidth} 
\centering
\includegraphics[height=3.5cm]{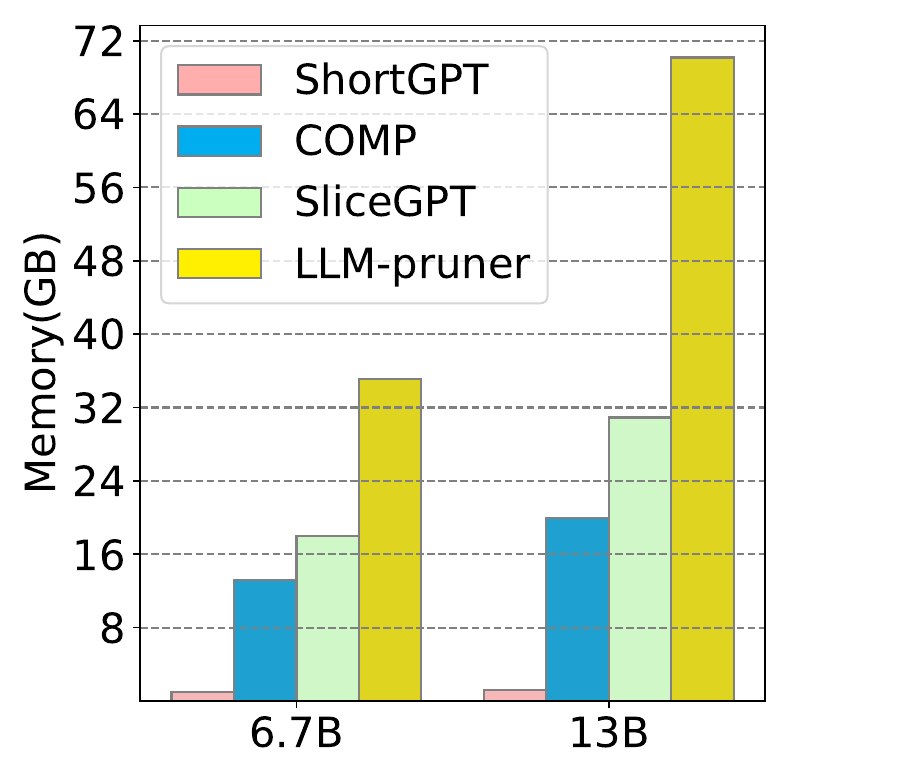}
\caption{OPT}
\label{fig:bar_OPT}
\end{subfigure}
\caption{GPU memory cost with different pruning methods.}
\label{fig:VRAMcost}
\end{figure}

\textbf{Memory Cost.}
To test the feasibility of COMP on resource-constrained devices, Figure \ref{fig:VRAMcost} illustrates the GPU memory overhead of different methods during pruning.
ShortGPT dynamically loads layers onto the GPU, 
thus the required memory for pruning is equivalent to that for a single-layer inference. 
COMP also dynamically loads layers during pruning, with the memory cost primarily arising from the evaluation of neuron importance.
By Figure \ref{fig:VRAMcost}, COMP only needs 8GB of memory to prune LLaMA-2-7B and 11GB of memory to prune LLaMA-2-13B.
For LLaMA-2-13B, COMP saves the memory overhead 62.68\% and 84.36\% compared with LLM-pruner and SliceGPT, respectively.
Note that compared to pruning LLaMA-2-7B, COMP consumes more memory (13GB) when pruning the OPT-6.7B model. 
That is because the maximum dense input dimension in OPT-6.7B is 16384, which is larger than that in LLaMA-2-7B (11008). 
This determines the dimensions of the coefficient matrix When evaluating the neuron importance. So, COMP is more memory-friendly for models with lower embedding dimensions.

\begin{figure}[t]
\centering
\includegraphics[height=3.6cm]{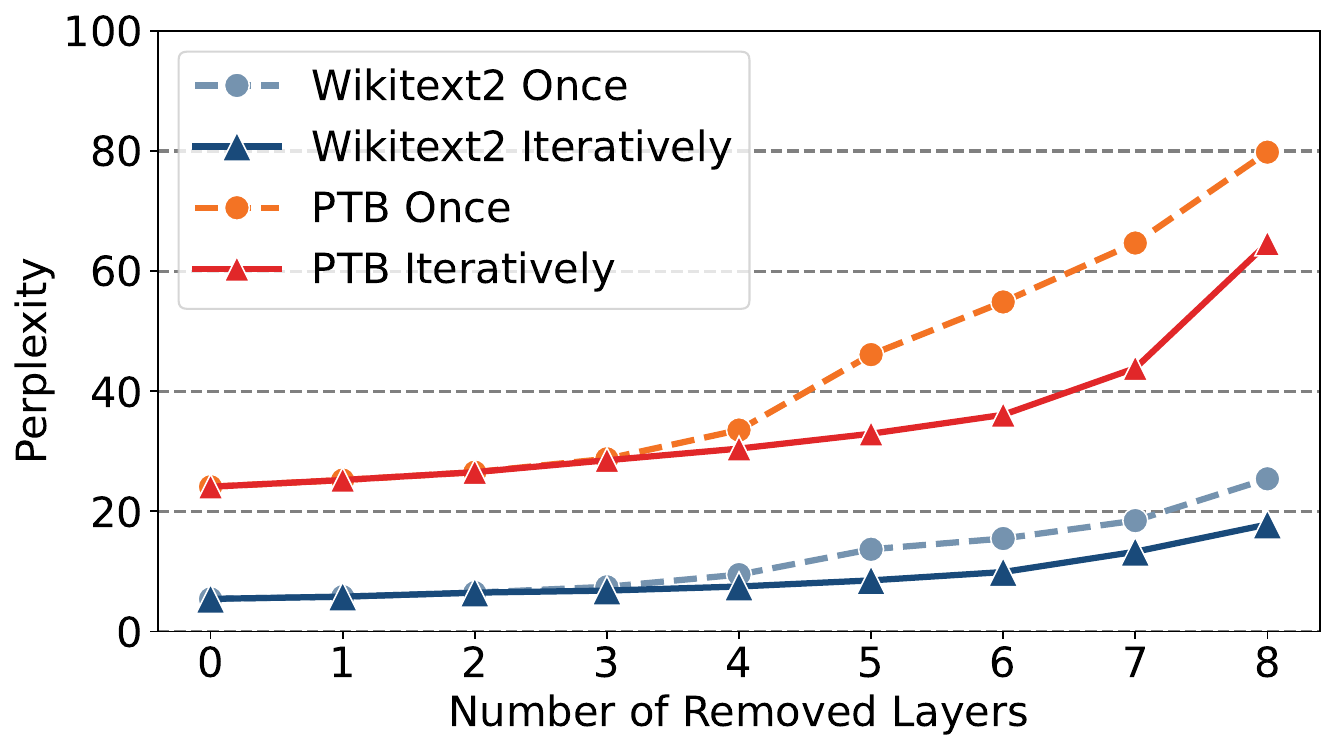}
\caption{The perplexity variation with the numbers of removed layers in LLaMA-2-7B.}
\label{fig:Diforder}
\end{figure}

\subsection{Ablation Study}
\textbf{Efficiency of Iterative Layer Order.}
We conduct a comparative analysis for the efficiency of calculating layer importance iteratively in section \ref{Layer-grained Pruning} and adopt calculating layer importance once as the baseline.
The perplexity results on the LLaMA-2-7B are shown in Figure \ref{fig:Diforder}.
The two calculation methods yield the same layer importance ranking when fewer than three layers are removed. 
However, when more than three layers are removed, the iterative method demonstrates a distinct advantage.
For example, when both remove 6 layers, the perplexity of the model with non-iterative layer removal is almost twice that of the model with iterative layer removal, which fully demonstrates the necessity of the iterative method.

\textbf{Efficiency of Identical Layer Input.}
In section \ref{Neuron-grained Pruning},  COMP used the same input as the original model when pruning neurons in each layer, rather than the input of the pruned layer. To verify the correctness of this approach, we tested two different methods on LLaMA-2-7B.
The perplexity results are shown in the Table \ref{table:cstlayeript}.
Obviously, using the same input is beneficial.
In addition, when the pruning ratio increases from 20\% to 30\%, the method that does not use the same input deteriorates more significantly.
Compared to the perplexity of using identical layer input, the perplexity that does not use identical input increases by 37.34\%, 70.72\% and 86.23\% respectively.
This is because when the pruning ratio increases, the output of each layer changes more than the output of the layer in the original model.
While the accumulated layer input shifts, the fine-tuned mask is overfitted on the calibration data.

\begin{table}[h]
\centering
\vspace{-0.1cm}
\caption{Wikitext2 perplexity of LLaMA-2-7B.}
\label{table:cstlayeript}
\adjustbox{max width=.4\textwidth}{
\begin{tabular}{cccc}
\toprule
Pruing ratio & 20\% &  25\% &  30\% \\
\midrule
\textbf{w/o} Identical inputs & 14.27 & 22.86 & 36.52 \\
\midrule
\textbf{w/} Identical inputs & \textbf{10.39} & \textbf{13.39} & \textbf{19.61} \\
\bottomrule
\end{tabular}}
\vspace{-0.3cm}
\end{table}

%% file: contents/conclusion.tex
In this paper, we propose a lightweight post-training structured pruning method COMP.
COMP first evaluates the layers' importance based on their input and output, and performs layer-grained Pruning.
Then, COMP executes neuron-grained pruning prunes in the dense of each remaining layer. 
Concretely, COMP sorts the importance of the input neurons in a dense based on the condition number, iteratively prunes neurons, and uses the mask tuning method to reconstruct the output of the pruned dense. 
COMP has low memory overhead and wide applicability, which is suitable for pruning LLMs on resource-constrained devices, requiring only a minimal amount of calibration data.

\textbf{Limitation} 
At a 30\% pruning ratio, COMP takes about 30 minutes to obtain the pruning results of LLaMA-2-7B in Table \ref{tab:LLaMA-2 performance}, and about 1 hour to obtain the results of LLaMA-2-13B.
When pruning the linear layer, COMP needs to iterate the pruning and mask adjustment steps to determine the number of pruned neurons, which is time-consuming.